\documentclass[letterpaper]{article} 
\usepackage[draft]{aaai2026}  
\usepackage{float}
\usepackage{bm}
\usepackage{times}  
\usepackage{helvet}  
\usepackage{courier}  
\usepackage[hyphens]{url}  
\usepackage{graphicx} 
\usepackage{natbib}  
\usepackage{caption} 
\usepackage{newfloat}
\usepackage{listings}
\DeclareCaptionStyle{ruled}{labelfont=normalfont,labelsep=colon,strut=off} 
\floatstyle{ruled}
\newfloat{listing}{tb}{lst}{}
\floatname{listing}{Listing}
\usepackage{colortbl}
\usepackage{amsmath}
\usepackage{amssymb}
\usepackage{mathtools}
\usepackage{amsthm}

\usepackage{booktabs}
\usepackage[ruled,vlined]{algorithm2e}
\usepackage{enumitem}
\usepackage{caption,tabularx,booktabs}
\usepackage{multicol}
\usepackage{multirow}
\usepackage{arydshln}
\usepackage{enumitem}
\usepackage{tabularray}

\title{Compressing KV Cache for Long-Context LLM Inference \\with Inter-Layer Attention Similarity}
\author{
    Da Ma\textsuperscript{\rm 1},
    Lu Chen\textsuperscript{\rm 1*},
    Situo Zhang\textsuperscript{\rm 1},
    Yuxun Miao\textsuperscript{\rm 1},
    Su Zhu\textsuperscript{\rm 2}
    Zhi Chen\textsuperscript{\rm 2}
    Hongshen Xu\textsuperscript{\rm 1}\\
    Hanqi Li\textsuperscript{\rm 1},
    Shuai Fan\textsuperscript{\rm 3},
    Lei Pan\textsuperscript{\rm 3},
    Kai Yu\textsuperscript{\rm 1*}
}
\affiliations{
    \textsuperscript{\rm 1}X-LANCE Lab, Department of Computer Science and Engineering \\
  MoE Key Lab of Artificial Intelligence, SJTU AI Institute \\
  Shanghai Jiao Tong University, Shanghai, China \\
  \textsuperscript{\rm 2}ByteDance
  \textsuperscript{\rm 3}AISpeech Co., Ltd., Suzhou, China\\
  \texttt{mada123@sjtu.edu.cn}

}

\usepackage[capitalize,noabbrev]{cleveref}
\crefname{section}{§}{§§}
\Crefname{section}{§}{§§}
\begin{document}

\maketitle

\renewcommand{\thefootnote}{\fnsymbol{footnote}}
\footnotetext[1]{*Corresponding authors.}
\thispagestyle{plain}
\pagestyle{plain}
\begin{abstract}
    The rapid expansion of context window sizes in Large Language Models~(LLMs) has enabled them to tackle increasingly complex tasks involving lengthy documents. However, this progress comes at the cost of a substantial increase in memory usage during inference, primarily due to the linear growth of the key-value~(KV) cache. Existing KV cache compression methods often discard less relevant tokens, which can lead to significant performance degradation when critical information is lost. In this paper, we propose \textsc{PoD}~(Proximal tokens over Distant tokens), a novel KV cache compression framework that allocates memory according to token importance, retaining less important tokens in a more compact, shared form rather than discarding them entirely. Our approach is motivated by two key observations: (1) proximal tokens---those at the beginning and end of the context---are significantly more important for next-token prediction, and (2) attention scores for distant tokens are highly redundant across consecutive layers. Leveraging these insights, \textsc{PoD} preserves the full KV cache for proximal tokens, while for distant tokens, it shares key states across layers. Since attention scores are determined by both queries and keys, sharing key states enables multiple layers to reuse a single set of keys for distant tokens, substantially reducing KV cache memory without discarding essential context. We further introduce a lightweight post-training adaptation to enable the model to adjust to this new attention-sharing structure. Extensive experiments on both synthetic~(Needle in a Haystack) and real-world long-context benchmarks demonstrate that \textsc{PoD} reduces KV cache memory usage by up to 35\% without compromising performance. Our method is orthogonal to existing token-selection-based techniques and can be combined with them for further KV cache compression.
\end{abstract}
\section{Introduction}
\label{sec:introduction}

Recently, the increasing context window size in Large Language Models~(LLMs)~\citep{NEURIPS2020_1457c0d6,achiam2023gpt,team2023gemini,reid2024gemini,touvron2023llama,touvron2023llama2,dubey2024llama} has allowed them to handle complex tasks requiring in-depth exploration of lengthy texts~\citep{bairi2024codeplan,mazumder2024lifelong}. However, it poses challenges to the memory footprint during inference. Specifically, since most LLMs are based on the Transformer~\citep{NIPS2017_3f5ee243} architecture, the size of key-value~(KV) cache~\citep{pope2023efficiently}, a widely used technique designed to prevent redundant computations, grows linearly with the context window size. Hence, compressing the KV cache during inference has become a critical problem for deploying LLMs with long context windows.

\begin{figure*}
    \centering
    \includegraphics[width=\linewidth]{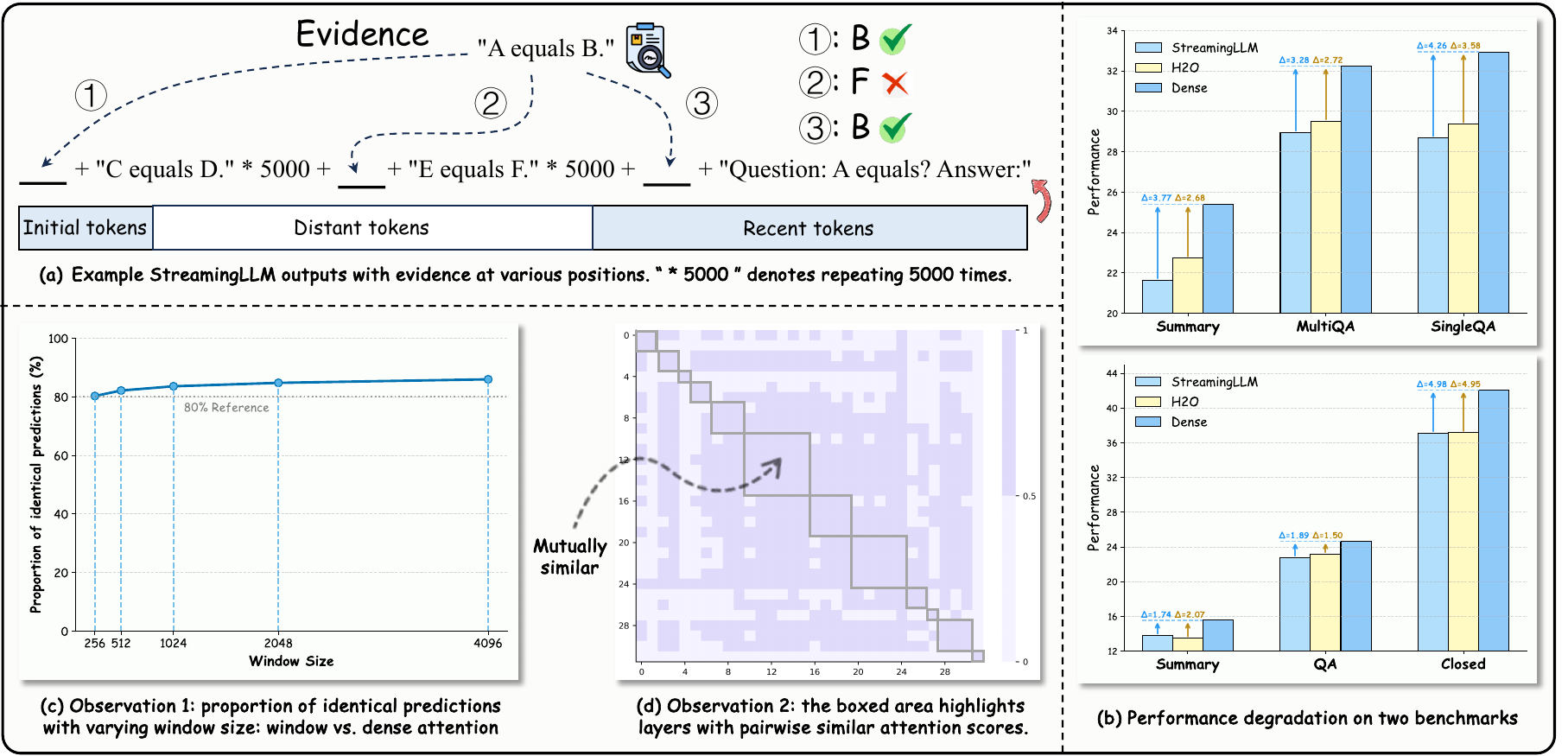}
    \caption{Experimental results from the LLaMA3-8B-32K model, including: (a) prediction failure example, (b) benchmark performance~(Details are in \cref{subsec:exp.perf}.), (c) window size impact on prediction consistency~(Details are in Appendix~\ref{subsec:obs1}.), and (d) attention similarity between layers~(Details are in Appendix~\ref{subsec:obs2}.).
    }
    \label{fig:intro}
\end{figure*}

Against this backdrop, recent studies have explored compressing the KV cache during inference by discarding less relevant tokens from the context. For example, window attention~\citep{Beltagy2020Longformer} keeps only the most recent tokens, while methods such as LM-Infinite~\citep{han-etal-2024-lm}, StreamingLLM~\citep{xiao2024efficient} and $\text{H}_2\text{O}$~\citep{zhang2023ho} further refine which tokens to retain based on their positional or contextual importance.  Although these approaches effectively reduce memory usage, they share a common drawback~\citep{tang2024quest}: \emph{critical tokens needed for subsequent text generation may be prematurely discarded}, leading to significant performance degradation when important information falls outside the cache. As shown in Figure \ref{fig:intro}-(a), when the important tokens~(evidence in the example) fall outside the cache~(the blue segment), the prediction fails. This limitation is further evidenced by the performance degradation of StreamingLLM and $\text{H}_2\text{O}$ on two real-world benchmarks~(see Figure \ref{fig:intro}-(b)).

In this paper, rather than simply discarding tokens to compress the KV cache, we propose a more nuanced approach that aims to minimize performance degradation while reducing memory usage. Our core motivation is that \emph{less important tokens should occupy less space in the KV cache, rather than being discarded entirely}. This perspective moves beyond hard pruning to retaining less important tokens in a more compact form, which raises two key questions: 1) how to identify, in a probabilistic sense, which tokens are more important for future generation, and 2) how to store less important tokens in a more compact form within the KV cache.

To address these questions, we examine two key properties of LLMs in long-context scenarios:
\begin{itemize}
    \item \textbf{Observation 1}: \emph{Proximal tokens~(i.e., initial and recent tokens) are, in a probabilistic sense, substantially more important for next-token prediction than distant tokens.}
Our empirical analysis on modern LLMs quantifies this: for $80\%$ of input positions, attending only to the $256$ nearest tokens leads to the same prediction as attending to the full context~(see Figure~\ref{fig:intro}-(c)).
    \item \textbf{Observation 2}: \emph{Attention scores between consecutive layers are highly similar.} While this phenomenon has been reported in smaller models~\citep{ijcai2019p735,bhojanapalli2021leveraging}, we show that it also holds for modern LLMs. As illustrated in Figure~\ref{fig:intro}-(d), attention scores for distant tokens remain strongly correlated between adjacent layers~(see the gray box), indicating substantial inter-layer redundancy.
\end{itemize}

Building on these two observations, we propose \textsc{PoD}~(\underline{P}roximal tokens \underline{o}ver \underline{D}istant tokens) for substantial KV cache compression in long-context LLMs. Motivated by the greater importance of proximal tokens~(Observation 1) and the inter-layer redundancy of attention scores for distant tokens~(Observation 2), \textsc{PoD} preserves the full KV cache for proximal tokens, while sharing attention scores across layers for distant tokens. As attention scores are determined by query and key states, this sharing enables multiple layers to reuse a single set of key states for distant tokens, substantially reducing KV cache memory without discarding essential context. Based on these principles, \textsc{PoD} operates in two main stages: 1) \emph{Exploration of Offline Inter-Layer Attention Sharing}~(\cref{subsec:model.asd}): determining which layers are suitable for sharing attention scores; 2) \emph{Lightweight Training Adaptation}~(\cref{subsec:model.training}): post-training the model on a limited dataset to adapt to the identified attention sharing patterns.

In addition, we conducted extensive experiments on Needle in a Haystack and three real-world long-context benchmarks. Results show that \textsc{PoD} reduces KV cache memory usage by up to 35\% without compromising model performance~(\cref{sec:experiments}). In summary, our main contributions are: 1) proposing a new KV cache compression paradigm that allocates memory based on token importance instead of discarding less important tokens; 2) developing \textsc{PoD}, which compresses the KV cache by sharing key states for distant tokens across layers, leveraging inter-layer attention redundancy; and 3) demonstrating that \textsc{PoD} achieves up to 35\% KV cache memory reduction with no performance loss. We will open-source our code and models.

\section{Methodology}
\label{sec:model}
Our approach (\textsc{PoD}) consists of two main steps: (1) grouping consecutive layers into blocks based on attention similarity analysis, and (2) sharing key states for distant tokens within each block, followed by lightweight post-training adaptation. We describe each step in detail below.

\begin{figure*}
    \centering
    \includegraphics[width=\linewidth]{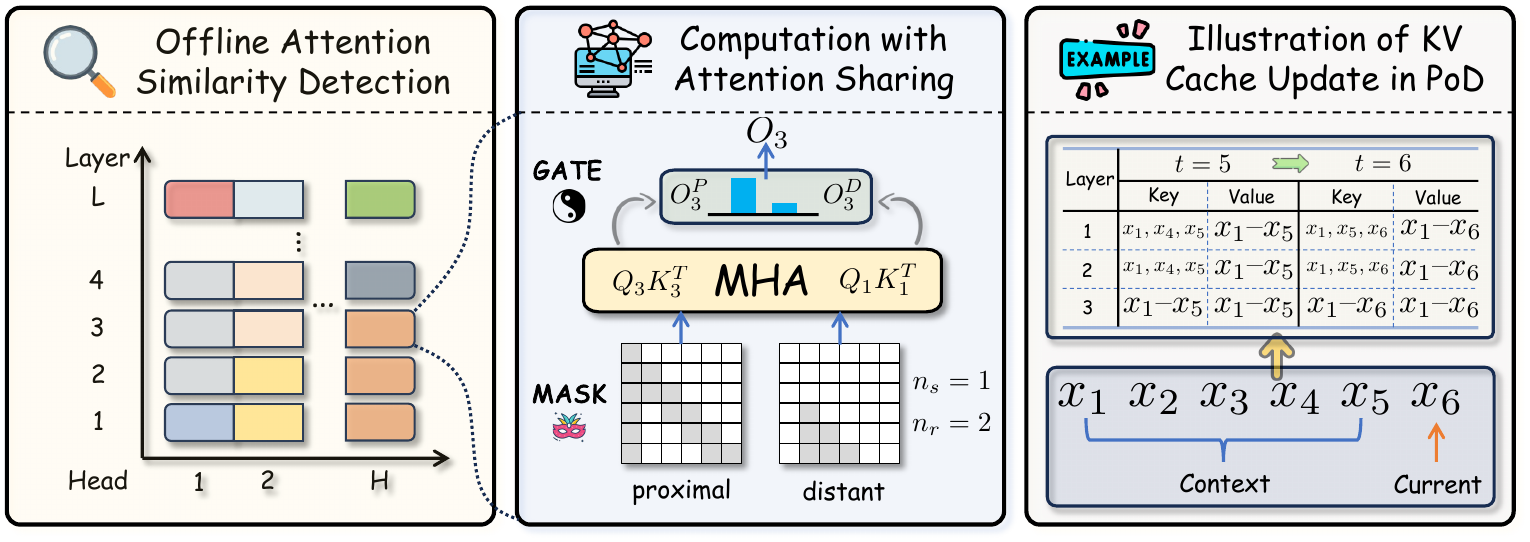}
    \caption{Overview of the \textsc{PoD} framework. \textbf{Left}: Example of head-wise layer partitioning based on inter-layer attention similarity; \textbf{Middle}: Key states for distant tokens are shared across layers within each block to reduce KV cache memory; \textbf{Right}: Example of KV cache update in \textsc{PoD}.}
    \label{fig:method}
\end{figure*}

\subsection{Offline Inter-Layer Attention Sharing Exploration}
\label{subsec:model.asd}
To guide the application of attention sharing, we first analyze the similarity of attention scores between layers in the LLM. This analysis enables us to group consecutive layers with similar attention patterns into blocks, which serve as the foundation for our sharing strategy.

\paragraph{Attention Score Calculation} 
Given $N$ input sequences $\big\{\mathbf{s}_i = (x_1, x_2, \ldots, x_n)\big\}_{i=1}^N$, we feed each sequence into the model $\mathcal{M}$ and extract the attention scores for the last $q$ tokens ($1 \leq q \leq n$) at every layer and attention head. Formally, for each sample, we obtain
\begin{align}
    \left\{\mathbf{S}_i^{\ell, h}\right\}_{1 \leq \ell \leq L,\, 1 \leq h \leq H} = \mathcal{M}(\mathbf{s}_i),
\end{align}
where $L$ and $H$ denote the number of layers and attention heads, respectively, and $\mathbf{S}_i^{\ell, h} \in \mathbb{R}^{q \times n}$ represents the attention scores at layer $\ell$ and head $h$ for the $i$-th input.

\paragraph{Attention Similarity Measurement}
To quantify the similarity between any two layers $\ell_a$ and $\ell_b$ ($1 \leq \ell_a,\, \ell_b \leq L$, $\ell_a \neq \ell_b$) for a given head $h$, we compute the average Jensen-Shannon (JS) divergence~\citep{MENENDEZ1997307} between their attention score distributions over the last $q$ tokens, aggregated across all $N$ samples:
\begin{align}
    sim_h(\ell_a, \ell_b) = \frac{1}{Nq} \sum_{i=1}^N \sum_{j=1}^q \text{JS}\left(\mathbf{S}_{i, j}^{\ell_a, h},\, \mathbf{S}_{i, j}^{\ell_b, h}\right),
\end{align}
where $\mathbf{S}_{i, j}^{\ell, h}$ denotes the $j$-th row of $\mathbf{S}_i^{\ell, h}$, corresponding to the attention distribution of the $j$-th token in the $i$-th input. The similarity score $sim_h(\ell_a, \ell_b)$ ranges from $0$ (completely dissimilar) to $1$ (identical).

\paragraph{Layer Grouping Strategy}
Based on the computed head-wise attention similarities, we group consecutive layers into blocks such that all layers within a block are sufficiently similar. Specifically, we consider two layers to be similar if $sim_h(\ell_a, \ell_b) \geq 0.5$. For each head, we employ a bottom-up greedy algorithm to iteratively merge consecutive similar layers into blocks, as detailed in Algorithm~\ref{alg:layer_grouping}. The resulting head-wise layer partitioning is exemplified in Figure~\ref{fig:method}~(left).

\subsection{Lightweight Training Adaptation}
\label{subsec:model.training}
To ensure the model can adapt to the new attention sharing mechanism within each block, we perform a lightweight post-training adaptation.

\paragraph{Attention Sharing within Each Block}
Let $\mathbf{s} = (x_1, x_2, \ldots, x_n)$ denote a long input sequence. In standard autoregressive Transformer-based LLMs, each token $x_i$ ($1 \leq i \leq n$) at layer $\ell$ attends to all previous tokens $\{x_j\}_{j \leq i}$. To reduce the memory and computation costs associated with distant tokens, we divide the preceding tokens into two groups: proximal tokens and distant tokens. Following prior works~\citep{han-etal-2024-lm,xiao2024efficient}, we treat both the most recent tokens and several initial tokens as proximal tokens, accounting for the ``attention sink'' phenomenon. Each token $x_i$ attends to both groups; however, for distant tokens, all layers within a block share the attention scores computed at the lowest layer of the block identified in Section~\ref{subsec:model.asd}.

Mathematically, for any attention head, let $\mathbf{Q}_{\ell},\, \mathbf{K}_{\ell},\, \mathbf{V}_{\ell} \in \mathbb{R}^{n \times d}$ denote the query, key, and value matrices at the $\ell$-th layer, respectively\footnote{For simplicity, we omit the attention head subscripts.}. Suppose layer $\ell$ belongs to block $B_\ell = \{\bar{\ell} \mid \ell_a \leq \bar{\ell} \leq \ell_b\}$, which consists of consecutive layers. For a given token $x_i$, we partition the preceding tokens into proximal and distant groups as described above. The attention outputs for $x_i$ with respect to proximal and distant tokens are computed as follows\footnote{If there are no distant tokens for $x_i$, the corresponding attention term is omitted.}:
\begin{equation}
    \label{formula:ie_attn}
    \begin{aligned}
    \mathbf{a}_{\ell, i}^P &= \frac{\mathbf{Q}_{\ell, i}\left[\mathbf{K}_{\ell, [1, n_s]};\, \mathbf{K}_{\ell, [n - n_r + 1, n]}\right]^T}{\sqrt{d}},\\
    \mathbf{o}_{\ell, i}^P &= \text{Softmax}\left(\mathbf{a}_{\ell, i}^P\right)\left[\mathbf{V}_{\ell, [1, n_s]};\, \mathbf{V}_{\ell, [n - n_r + 1, n]}\right],\\
    \mathbf{a}_{\ell, i}^D &= \frac{\mathbf{Q}_{\ell_a, i}\, \mathbf{K}_{\ell_a, [n_s + 1, n - n_r]}^T}{\sqrt{d}},\\
    \mathbf{o}_{\ell, i}^D &= \text{Softmax}\left(\mathbf{a}_{\ell, i}^D\right)\, \mathbf{V}_{\ell, [n_s + 1, n - n_r]},
    \end{aligned}
\end{equation}
where $\mathbf{Q}_{\ell, i}$ denotes the $i$-th row of $\mathbf{Q}_{\ell}$, and $\mathbf{K}_{\ell, [a, b]}$ denotes the rows from $a$ to $b$ (inclusive) of $\mathbf{K}_\ell$. Here, $n_s$ (start size) and $n_r$ (recent size) represent the number of initial and most recent tokens classified as proximal tokens, respectively, and $[\cdot;\cdot]$ denotes concatenation. $\mathbf{a}_{\ell, i}^P \in \mathbb{R}^{1 \times (n_s + n_r)}$ and $\mathbf{o}_{\ell, i}^P \in \mathbb{R}^{1 \times d}$ are the attention logits and outputs for proximal tokens, respectively; analogous notations apply to distant tokens.



\paragraph{Aggregation of Attention Outputs to Proximal and Distant Tokens} 
To combine the attention outputs from proximal and distant tokens, we employ a parameter-free gating mechanism\footnote{The derivation of the gating formula is provided in Appendix~\ref{subsec:appendix_derivation}.}:
\begin{equation}
\label{formula:gate}
    \begin{aligned}
    g_{\ell, i} &= \frac{\sum \exp{\mathbf{a}_{\ell, i}^P}}{\sum \exp{\mathbf{a}_{\ell, i}^P} + \sum \exp{\mathbf{a}_{\ell, i}^D}},\\
    \mathbf{o}_{\ell, i} &= g_{\ell, i} \cdot \mathbf{o}_{\ell, i}^P + (1 - g_{\ell, i}) \cdot \mathbf{o}_{\ell, i}^D,
    \end{aligned}
\end{equation}
where $g_{\ell, i}$ adaptively balances the contributions from proximal and distant tokens for each token $x_i$ at layer $\ell$.
Figure~\ref{fig:method}~(middle) provides a detailed example of the computation under the attention sharing scheme, highlighting the attention masks applied to proximal and distant tokens.

\begin{algorithm}[t]
\caption{Greedy Layer Grouping Algorithm}
\label{alg:layer_grouping}
\KwIn{Head-wise attention similarities between layers: $\left\{sim_h\left(\ell_a, \ell_b\right)\right\}_{1\leq \ell_a,\,\ell_b\leq L}^{1\leq h \leq H}$} 
\KwOut{Head-wise layer blocks}

head\_wise\_layer\_blocks $\leftarrow \left[\,\right]$\;

\For{{\upshape head} $h\leftarrow 1$ to $H$} {
    current\_head\_layer\_blocks$\leftarrow$ $\left[\left\{1\right\}\right]$\tcp*{Each block is a set.}
    \For{{\upshape layer} $\ell\leftarrow 2$ to $L$} {
        current\_block$\leftarrow$ the last element of current\_head\_layer\_blocks\;
        \tcp{Layer $\ell$ is similar to all layers in the current block.}
        \If{$sim_h\left(\ell, \hat{\ell}\right)\geq 0.5,\forall\hat{\ell}\in$ \text{\upshape current\_block}} { 
            Add $\ell$ to current\_block\;
        }
        \Else {
            Append $\left\{\ell\right\}$ to current\_head\_layer\_blocks\;
        }
    }
    Append current\_head\_layer\_blocks to head\_wise\_layer\_blocks\;
}
\textbf{Return} head\_wise\_layer\_blocks\;
\end{algorithm}

\section{Experiments}
\label{sec:experiments}

In this section, we mainly address two key questions:
\begin{itemize}
    \item Does \textsc{PoD} maintain model performance in long-context scenarios?
    \item Can \textsc{PoD} effectively reduce KV cache memory usage during long-context inference?
\end{itemize}

\paragraph{Implementation Details}
For data preparation, we sampled $5$B tokens from Dolma~\citep{soldaini-etal-2024-dolma} for post-training, ensuring that the number of tokens in each sequence length interval remains consistent~\citep{glm2024chatglm}. 

The baseline model, LLaMA3-8B-32K, was obtained by post-training LLaMA3-8B on $5$B tokens with a maximum sequence length of $32$K.
To construct the \textsc{PoD} model, we first performed offline attention similarity detection on LLaMA3-8B-32K~(see Appendix~\ref{subsec:obs2} for details) to determine the optimal layer grouping for attention sharing, which resulted in approximately $35\%$ KV cache memory savings.
Subsequently, we continued post-training from LLaMA3-8B-32K on the same $5$B tokens~(with a maximum sequence length of $32$K), applying the new attention sharing structure with $n_s = 16$ and $n_r = 4080$, to adapt the model to these architectural changes.

During training, we used a batch size of $4$M tokens and set the learning rate to $1\mathrm{e}{-5}$ with a cosine annealing scheduler. The RoPE (Rotary Positional Embedding)~\citep{su2023roformerenhancedtransformerrotary} base was set to 16M+, following the approach of~\cite{xiong-etal-2024-effective}.
For implementation, we adopted the HuggingFace~\citep{wolf-etal-2020-transformers} and DeepSpeed~\citep{10.1145/3394486.3406703} frameworks, incorporating ZeRO-3~\citep{10.5555/3433701.3433727} and Ulysses~\citep{jacobs2023deepspeed} sequence parallelism. Additionally, we employed the efficient FlexAttention module from PyTorch~\citep{paszke2019pytorch}.

\paragraph{Baselines} We consider three types of baselines:
\begin{itemize}
    \item \emph{Token-selection-based methods}: {\tt SnapKV}~\citep{li2024snapkv} selects and caches important tokens based on attention scores during prefilling. {\tt PyramidKV}~\citep{zhang2024pyramidkv} extends SnapKV by varying the number of cached tokens across layers. {\tt Quest}~\citep{tang2024quest} does not reduce KV cache size, but decreases the number of tokens involved in attention computation via efficient token selection.
    \item \emph{Token-eviction-based methods}: {Window Attention~({\tt WA})}~\citep{Beltagy2020Longformer} restricts each token to attend only to a local window. {\tt WA+CPT} further post-trains the model with window attention. {\tt StreamingLLM}~\citep{xiao2024efficient} and {\tt LM-Infinite}~\citep{han-etal-2024-lm} allow tokens to attend to both neighboring and initial tokens, with LM-Infinite using different position embeddings. {\tt {$\text{H}_2\text{O}$}} dynamically adds or removes tokens based on attention scores during decoding.
    \item \emph{Layer-sharing-based methods}: {\tt CLA}~\citep{brandon2024reducing} reduces KV cache by sharing key and value states across adjacent layers.
\end{itemize}
More details are in Appendix~\ref{sec:appendix.baseline}.

\begin{table*}
\centering

\renewcommand{\arraystretch}{1.1}
\setlength{\tabcolsep}{1.8mm}
\begin{tabular}{lc|ccccc;{1pt/1pt}c|ccc;{1pt/1pt}c} 
\hline

\hline
\multirow{2}{*}{\textbf{Model}} & \multirow{2}{*}{\textbf{Window}} & \multicolumn{6}{c|}{\textbf{\textit{LongBench}}} & \multicolumn{4}{c}{\textbf{\textit{LEval}}}                                                                        \\ 
\cdashline{3-12}[1pt/1pt]
                  &                   & \textit{SQA}  & \textit{MQA}  & \textit{Summ}  & \textit{Few-Shot} & \textit{Code} &   \textbf{\textit{Avg.}}     & \multicolumn{1}{c}{\textit{Closed}} & \multicolumn{1}{c}{\textit{QA}} & \multicolumn{1}{c;{1pt/1pt}}{\textit{Summ}} & \multicolumn{1}{c}{\textbf{\textit{Avg.}}}  \\ 
\hline
{\tt LLaMA3-8B-32K} & \text{32K}  & $32.9$  & $32.2$ & $25.4$ & $69.3$ & $66.5$ & $45.3$ & $42.1$ & $24.7$ & $15.6$ & $27.5$     \\
\hline

\multicolumn{12}{c}{\textbf{\emph{Token-selection-based methods}}}\\
\hline

{\tt SnapKV} & \text{4K} & $31.8$&	$31.9$&	$21.9$&	$68.6	$&$66.7	$&$44.2$&	$39.9	$&$23.9$&	$13.5$&	$25.8$ \\
{\tt PyramidKV} & \text{4K} & $33.3$&	$31.5$&	$23.8$&	$68.9$&	$66.4$&	$44.8$	&$42.1$&	$22.6$&	$13.0$&	$25.9$ \\
{\tt Quest}  & \text{4K} & $32.1$&	$32.2$&	$24.3$&	$69.1$&	$66.4$&	$44.8$&	$40.6$&	$25.6$&	$14.7$&	$26.9$\\
\hline

 \multicolumn{12}{c}{\textbf{\emph{Token-eviction-based methods}}}\\
\hline

{\tt LM-Infite}   &  \text{16+4080}     & $28.8$ & $29.0$ & $21.7$ & $68.1$ & $66.5$ & $42.8$  & $37.3$ & $22.8$  & $13.9$  & $24.7$      \\
{\tt StreamingLLM}   &  \text{16+4080}     & $28.7$  & $29.0$ & $21.6$ & $68.1$ & $66.6$ & $42.8$  & $37.1$  & $22.8$ & $13.8$  & $24.6$      \\
{\tt $\text{H}_2\text{O}$}   &  \text{96+4000}     & $29.4$  & $29.5$ & $22.7$ & $68.5$ & $66.2$ & $43.2$  & $37.2$ & $23.2$  & $13.5$  & $24.6$      \\
{\tt WA}   &  \text{4K}     & $8.9$ & $3.6$ & $9.1$ & $11.1$ & $41.1$ & $14.8$  & $21.0$ &  $5.6$ & $2.8$  &  $9.8$     \\
{\tt WA+CPT}   &  \text{4K}     & $26.9$ & $28.0$ & $22.3$ & $66.6$ & $66.1$ &  $42.0$ & $32.9$ &  $22.1$ &  $12.6$ & $22.5$      \\
\hline

\multicolumn{12}{c}{\textbf{\emph{Layer-sharing-based methods}}}\\
\hline
{\tt CLA} & $\text{32K}$ & $24.0$&	$22.6$&	$22.5$&	$60.9$&	$59.4$&	$37.9$&	$19.1$&	$13.5$&	$11.5$&	$14.7$\\
\cdashline{1-12} 
{\tt \textsc{PoD}}~(ours)   &  \text{16+4080+28K}     & $31.0$ & $32.4$ & $24.8$ & $67.3$ & $68.3$ & $44.8$  & $43.6$ &  $23.0$ & $15.0$  &  $27.2$     \\
{\tt \textsc{PoD}+\text{SnapKV}}~(ours) & $\text{4K}$ & $31.0$ & $32.7$ &	$22.9$ & $66.9$ &	$67.8$ & $44.3$ & $43.1$ & $22.1$ & $14.3$ & $26.5$ \\

\hline

\hline
\end{tabular}

\caption{Evaluation results of different methods on two famous long context benchmarks}
\label{tab:benchmark}
\end{table*}

\begin{table}
\centering

\renewcommand{\arraystretch}{1.05}
\setlength{\tabcolsep}{5.5mm}
\begin{tabular}{lc} 
\hline

\hline
\textbf{Method} & \textbf{Score~(\%)} \\ 

\hline

\hline
{\tt Dense} & $97.9$ \\
\hline
{\tt SnapKV} & $98.3$ \\
{\tt PyramidKV} & $97.8$ \\
{\tt StreamingLLM} & $56.8$\\
{\tt $\text{H}_2\text{O}$} & $55.6$\\
{\tt CLA} & $64.8$\\
\hline
{\tt \textsc{PoD}}~(ours) & $98.9$\\
{\tt \textsc{PoD}+SnapKV}~(ours) & $94.6$\\

\hline

\hline
\end{tabular}

\caption{Scores for needle in a haystack}
\label{tab:needle}
\end{table}

\subsection{Performance Evaluation}
\label{subsec:exp.perf}

To evaluate the performance of \textsc{PoD}, we conducted experiments in two fields: 1) Needle in a Haystack and 2) Practical Long Context Benchmarks.

\paragraph{Needle in a Haystack} 
Table~\ref{tab:needle} presents a quantitative comparison of different methods on the needle-in-a-haystack task, where a random statement~(the ``needle'') is placed in the middle of a long context window and the model is asked to retrieve it. The results show that \textsc{PoD} and other token-selection-based methods~(e.g., SnapKV, PyramidKV) achieve near-perfect performance, significantly outperforming token-eviction-based methods~(StreamingLLM, $\text{H}_2\text{O}$) and the layer-sharing-based method~(CLA). Notably, \textsc{PoD} and token-selection-based approaches are orthogonal and can be combined, as demonstrated by the strong performance of \textsc{PoD}+SnapKV. The visual illustration of the search results is shown in Appendix~\ref{subsec:visrniah}.


\paragraph{Long Context Benchmarks} To ensure that \textsc{PoD} can handle real-world tasks, we evaluated it on two well-known long context benchmarks: LongBench~(English version)~\citep{bai-etal-2024-longbench} and LEval~\citep{an-etal-2024-l}. We test on $14$ datasets within LongBench involving Single-document QA, Multi-document QA, Summarization, Few-shot learning, and Code completion tasks. LEval consists of $20$ sub-tasks, divided into two groups: closed-domain and open-domain. The closed-domain group primarily evaluates reasoning and comprehension over longer contexts, while the open-domain group focuses on tasks such as summarization and question answering, which require aggregating information from long documents.

\begin{table}[t]
\centering

\renewcommand{\arraystretch}{1.15}
\setlength{\tabcolsep}{1.8mm}
\begin{tabular}{c|cc;{2pt/2pt}ccc} 
\hline

\hline
\textbf{Theoretical} & \multicolumn{5}{c}{\textbf{Practical maximum batch size $b$}}  \\ 
\cdashline{2-6}[2pt/2pt]
          \textbf{saving}        & $x$ & $y$ & {\tt Dense} & {\tt \textsc{PoD}   }   &    $\bm{\uparrow} $    \\ 
\hline
        \multirow{4}{*}{$35\%$}          & $2048$  & $8192$ & $25$ & $33$&$32.0\%$             \\
        & $4096$ & $8192$ & $13$ & $17$ & $30.8\%$ \\
        & $8192$ & $8192$ & $6$ & $8$ & $33.3\%$ \\
        & $16384$ & $8192$ & $3$ & $4$ & $33.3\%$ \\
        
\hline

\hline
\end{tabular}

\caption{Theoretical and practical memory footprint savings }
\label{tab:memory}
\end{table}
\begin{table*}
\centering

\renewcommand{\arraystretch}{1.05}
\setlength{\tabcolsep}{1.2mm}
\begin{tabular}{l|ccccc;{1pt/1pt}ccccc;{1pt/1pt}ccccc|c} 
\hline

\hline
\multirow{2}{*}{\textbf{Model}} & \multicolumn{5}{c;{1pt/1pt}}{\textbf{\textit{32K}}} & \multicolumn{5}{c;{1pt/1pt}}{\textbf{\textit{64K}}} & \multicolumn{5}{c|}{\textbf{\textit{128K}}} & \multirow{2}{*}{\textbf{\textit{Avg.}}}                                                                        \\ 
& \textit{QA} & \textit{MC} & \textit{Summ} & \textit{PK} & \textit{NS} & \textit{QA} & \textit{MC} & \textit{Summ} & \textit{PK} & \textit{NS} & \textit{QA} & \textit{MC} & \textit{Summ} & \textit{PK} & \textit{NS} &\\
\hline
{\tt LLaMA3.1-8B}   & $29.6$ & $25.3$	&$15.5$	&$27.1$	&$27.1$& $37.3$	&$28.8$	&$15.1$	&$54.2$	&$54.2$ & $31.2$ 	&$41.8$	&$14.9$	&$100.0$	&$99.5$	&$40.1$     \\

\hline
\multicolumn{17}{c}{\textbf{\emph{Token-selection-based methods}}}\\
\hline
{\tt SnapKV}  & $28.9$ & $26.2$&	$13.3$&	$27.1$&	$26.6$& $35.1$ &	$28.0$&	$11.7$&	$54.2$&	$53.1$ & $29.3$ & $34.5$&	$13.7$&	$100.0$&	$99.0$&	$38.7$ \\
{\tt PyramidKV}  & $29.9$ & $26.2$&	$15.0$&	$27.1$&	$27.1$& $36.5$ &  $29.3$&	$15.2$&	$54.2$&	$54.2$&\multicolumn{5}{c|}{\text{OOM}\footnotemark[4]} & \text{OOM} \\
{\tt Quest}   & $28.5$& $25.8$&	$12.9$&	$27.1$&	$27.1$& $35.8$ &	$27.5$&	$11.2$&	$54.2$&	$54.2$& $29.0$ & $34.5$&	$9.0$&	$100.0$&	$98.1$&	$38.3$ \\

\hline
\multicolumn{17}{c}{\textbf{\emph{Token-eviction-based methods}}}\\
\hline

{\tt LM-Infite}       &  $25.3$&	$26.6$&	$12.5$&	$3.4	$&$3.4	$&$27.3$&	$28.8$&	$12.6$&	$3.4	$&$3.4	$&$22.2$&	$29.3$&	$13.0$&	$3.4	$&$3.2	$&$14.5$ \\
{\tt StreamingLLM}       & $25.4$&	$26.2$&	$12.1$&	$3.4	$&$3.4	$&$27.1$&	$28.0$&	$13.3$&	$3.4	$&$3.4	$&$22.7$&	$28.0$&	$12.8$&	$3.4	$&$3.2	$&$14.4$ \\
{\tt $\text{H}_2\text{O}$}      &  $25.4$&	$26.2$&	$13.1$&	$4.9	$&$3.4	$&$27.6$&	$28.0$&	$14.3$&	$7.1	$&$3.6 $&\multicolumn{5}{c|}{\text{OOM}\footnotemark[4]} & \text{OOM}\\
{\tt WA}       &   $3.5$&	$3.5$&	$0.5$&	$0.0$&	$1.4$&	$3.4$&	$3.1$&	$0.7$&	$0.0$&	$1.2$&	$3.3$&	$3.5$&	$0.7$&	$0.0$&	$1.2$&	$1.7$  \\
{\tt WA+CPT}     &  $12.6$&	$18.8$&	$11.3$&	$3.4	$&$3.4	$&$13.1$&	$17.9$&	$10.8$&	$3.4	$&$3.4	$&$12.8$&	$21.0$&	$11.3$&	$3.4	$&$3.4	$&$10.0$\\

\hline
\multicolumn{17}{c}{\textbf{\emph{Layer-sharing-based methods}}}\\
\hline
{\tt CLA}  & $22.1$&	$34.1$&	$13.6$&	$24.1$&	$25.8$&	$22.7$&	$31.9$&	$12.7$&	$50.9$&	$52.5$&	$21.6$&	$34.5$&	$13.0$&	$97.8$&	$96.8$&	$36.9$\\
{\tt \textsc{PoD}}       & $27.4$&	$35.4$&	$17.9$&	$26.6$&	$27.1$&	$29.6$&	$36.9$&	$15.5$&	$53.7$&	$54.2$&	$26.6$&	$42.8$&	$15.5$&	$99.8$&	$99.2$&	$40.6$\\
{\tt \textsc{PoD}+\text{SnapKV}} & $28.6$&	$35.9$&	$13.9$&	$26.6$&	$23.4$&	$28.5$&	$36.2$&	$14.0$&	$53.7$&	$49.5$&	$24.7$&	$40.6$&	$12.4$&	$99.8$&	$88.0$&	$38.4$ \\

\hline

\hline
\end{tabular}
\caption{Evaluation results on InfiniteBench~(128K). OOM: out of memory over one A800-80G GPU}
\label{tab:infinitebench}
\end{table*}

Table \ref{tab:benchmark} illustrates all experimental results. To ensure fairness, all baseline attention mechanisms have the same window size. For \textsc{PoD}, we also ensure that the number of proximal tokens each token attends to is consistent with this window size. We can draw the following conclusions: 1) \textsc{PoD} outperforms token-eviction-based methods, demonstrating that our approach of not losing tokens is indeed effective. 2) With a small amount of post-training data, \textsc{PoD} beats the classical layer-sharing-based method CLA, demonstrating that our model has an advantage in adapting existing LLMs. 3) 
Both PoD and token-selection-based methods can achieve performance comparable to the standard dense model. Furthermore, \textsc{PoD} is \emph{orthogonal} to token-selection-based methods, and combining them can further reduce the size of the KV cache while maintaining model performance.

\subsection{Efficiency Evaluation}
\paragraph{KV Cache Memory}
The savings in memory consumption can be analyzed from both theoretical and empirical perspectives. Theoretically, we can calculate the potential reduction in KV cache size based on the layer-sharing results obtained from offline analysis. Empirically, we can conduct end-to-end evaluations to assess the actual savings. Following FlexGen~\citep{10.5555/3618408.3619696} and LCKV~\citep{wu-tu-2024-layer}, for a prompt of length $x$, we let the model generate $y$ tokens, The maximum batch size $b$ achievable on a given GPU will be used to assess the memory requirements of the model. A larger $b$ indicates that the model is more memory-efficient. Table \ref{tab:memory} presents the memory consumption results. We observe that \textsc{PoD} achieves a more than $30\%$ increase in maximum batch size across varying input text lengths, closely aligning with our theoretical KV cache savings rate of $35\%$, demonstrating that \textsc{PoD} effectively reduces memory usage.

\paragraph{Latency and Efficiency} Table~\ref{tab:balance} compares various methods in terms of latency~(throughput), KV cache savings, and performance. Our analysis highlights clear trade-offs: 
\begin{itemize}[leftmargin=1em]
    \item Token-eviction-based methods offer fast throughput and high KV cache savings but incur notable performance loss.
    \item Token-selection-based methods compress the KV cache with minimal performance impact, but the token selection step slows down inference.
    \item \textsc{PoD} achieves lower latency than SnapKV while still maintaining strong performance and cache savings, and combining \textsc{PoD} with token-selection-based methods brings further gains across all dimensions.
\end{itemize}
However, all performance-preserving methods---such as SnapKV, PyramidKV, and \textsc{PoD}---still incur some latency overhead. Improving their efficiency is an important direction for future work.

\begin{table}[t]
\centering

\renewcommand{\arraystretch}{1.02}
\begin{tabular}{l|cccc} 
\hline

\hline
\multirow{2}{*}{\textbf{Method}} & \multirow{2}{*}{\textbf{Stage}} & \multirow{2}{*}{\textbf{Tp}~$\bm{\uparrow}$} & \textbf{KV}~$\bm{\uparrow}$  & \textbf{Perf}$~\bm{\downarrow}$ \\
&  &  &  (\%) & (\%)\\
\hline
{\tt Dense} & - & $40.3$ &- &  -\\
\hline
\multicolumn{5}{c}{\textbf{\emph{Token-selection-based methods}}}\\
\hline
{\tt SnapKV} & \multirow{2}{*}{P} & $24.5$ & $87.5$ &  $4.3$ \\
{\tt PyramidKV} & & $20.6$ &$93.6$ &  $3.4$ \\
\hline

\multicolumn{5}{c}{\textbf{\emph{Token-eviction-based methods}}}\\
\hline
{\tt LM-Infinite} & \multirow{5}{*}{P\&D} & $42.8$ & \multirow{5}{*}{$87.5$} &  $7.7$\\
{\tt StreamingLLM} & & $43.0$ &&  $8.0$\\
{\tt $\text{H}_2\text{O}$} & & $27.7$ &&  $7.4$ \\
{\tt WA} & & $50.1$ &&  $65.9$\\
{\tt WA+CPT} & &$50.1$ & &  $12.6$\\
\hline
\multicolumn{5}{c}{\textbf{\emph{Layer-sharing-based methods}}}\\
\hline
{\tt CLA} & \multirow{3}{*}{P\&D} & $41.8$ & $50.0$ &  $31.4$\\
{\tt \textsc{PoD}}~(ours) & & $31.7$ & $35.0$ & $2.8$ \\
{\tt \textsc{PoD}+SnapKV}~(ours) & & $23.4$ &$91.9$ &  $3.1$\\

\hline

\hline
\end{tabular}
\caption{Comparison of different methods from multiple perspectives. \textbf{Stage}: optimized stage.`P' indicates Prefilling and `D' indicates Decoding; \textbf{Tp}: throughput~(tokens per second); \textbf{KV}: KV cache saving; \textbf{Perf}: performance degradation.
}
\label{tab:balance}
\end{table}
\subsection{Additional Analysis}

\paragraph{Scaling to longer context and other LLMs} {To explore the generality of our method, we conducted experiments on LLaMA3.1-8B}~\citep{dubey2024llama}{, which can handle longer~($128$K) contexts. We sampled $5$B tokens from the ProLong-data-512K}~\citep{gao2024trainlongcontextlanguagemodels} {dataset and applied the same hyperparameter configuration used for training LLaMA3-8B-32K to post-train LLaMA3.1-8B with a sequence length of $128$K. The evaluation results over $5$ sub-tasks in the InfiniteBench}~\citep{zhang-etal-2024-bench} {under different context sizes are shown in Table} \ref{tab:infinitebench}. 

{Consistent with the conclusions found in Table} \ref{tab:benchmark}, {our method causes less performance degradation compared to token-eviction-based methods. However, a notable difference is that token-selection-based methods appear to struggle in maintaining model performance in longer context scenarios. This limitation is also reflected in the combined model~(\textsc{PoD}+SnapKV), which integrates our method with token-selection-based methods, showing a decline in performance. This to some extent indicates that our method is more robust to the context length.}

\begin{table}
\centering

\setlength{\tabcolsep}{1.7mm}
\begin{tabular}{cccccc|c} 
\hline

\hline
\textbf{Method} & \textbf{MMLU} & \textbf{HS} & \textbf{Arc-e} & \textbf{Arc-c} & \textbf{HE} & \textbf{Avg.}\\
\hline
{\tt Dense} &$61.4$ &	$58.0$ &	$80.9$ &	$70.2$ &	$28.0$ &	$59.7$ \\
{\tt CLA} &$36.1$ &	$52.6$ &	$23.0$ &	$28.4$ &	$14.6$ &	$30.9$ \\
\hline
{\tt \textsc{PoD}} & $62.8$ &	$59.7$ &	$83.5$ &	$73.6$ &	$29.3$ &	$61.8$
 \\
        
\hline

\hline
\end{tabular}
\caption{Results on standard benchmarks. HS and HE denote Hellaswag and HumanEval, respectively.}
\label{tab:normal}
\end{table}

\begin{figure}[t]
    \includegraphics[width=\linewidth]{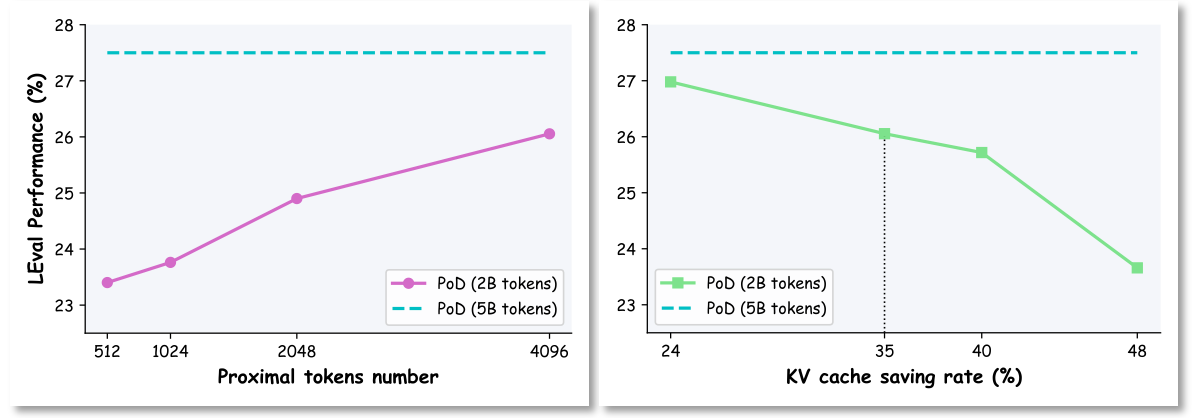}
    \caption{
    Ablation study on key hyperparameters in \textsc{PoD}. \textbf{Left:} LEval performance vs. proximal tokens number. 
\textbf{Right:} LEval performance vs. KV cache saving rate. 
$4$K proximal tokens and $35\%$ saving rate provide a good balance of accuracy and KV cache compression rate.
    }
    \label{fig:ablation}
\end{figure}

\paragraph{Ablation Study on Key Hyperparameters in \textsc{PoD}}
We examine how two key hyperparameters in \textsc{PoD}---the number of proximal tokens and the KV cache saving rate---affect model performance. Starting from LLaMA3-8B-32K, we continued training with $2$B tokens. As shown in Figure~\ref{fig:ablation}~(left), increasing the number of proximal tokens consistently improves performance, with $4$K proximal tokens enabling the model to match the performance of LLaMA3-8B-32K trained on $5$B tokens. Figure~\ref{fig:ablation}~(right) shows that higher KV cache saving rates lead to decreased performance. To strike a balance between KV cache compression and accuracy, we set the number of proximal tokens to $4$K and compressed the KV cache to $35\%$, maintaining competitive performance with reduced resource usage.

\paragraph{Evaluation on Standard Benchmarks} Table~\ref{tab:normal} shows model performance on standard benchmarks. \textsc{PoD} matches the dense baseline with no accuracy loss, while CLA suffers significant degradation, likely due to indiscriminate information sharing across layers.

\begin{figure}
    \includegraphics[width=\linewidth]{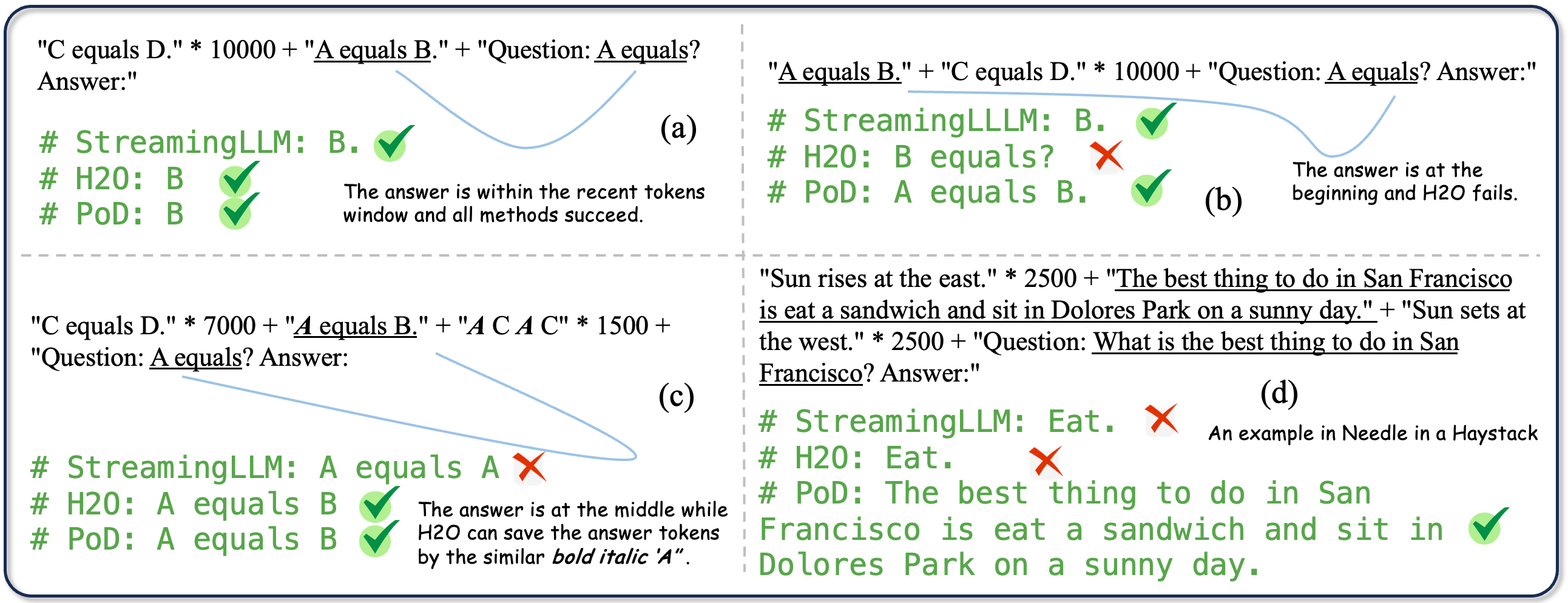}
    \caption{Case study of different methods. $s*n$ means repeating $n$ times of the string $s$. $+$ represents the concatenation of strings.}
    \label{fig:case}
\end{figure}

\paragraph{Case Study} 
Figure~\ref{fig:case} compares StreamingLLM, $\text{H}_2\text{O}$, and \textsc{PoD} on four representative cases. 
In case (a), the answer is within the window of recent tokens, so all methods make correct predictions. 
In case (b), the answer is at the beginning; StreamingLLM and \textsc{PoD} retain it and predict correctly, but $\text{H}_2\text{O}$ discards it due to many irrelevant tokens, resulting in an error. 
In case (c), the answer is in the middle; StreamingLLM cannot access it and fails, while $\text{H}_2\text{O}$ and \textsc{PoD} both succeed. 
In case (d), only \textsc{PoD} can find the answer in the ``Needle in a Haystack" scenario, as the other methods overlook the answer tokens.

\section{Related Work}
\label{sec:related-work}

Long-context LLMs face significant memory challenges due to their large parameter sizes and lengthy input sequences. Existing optimization approaches for reducing KV cache memory can be broadly categorized into three areas.

\paragraph{Context Compression and Computation Optimization}
Many methods reduce memory usage by discarding less important tokens from the KV cache. For example, window attention~\citep{Beltagy2020Longformer} retains only the most recent tokens, while LM-Infinite~\citep{han-etal-2024-lm} and StreamingLLM~\citep{xiao2024efficient} preserve both initial and recent tokens. $\text{H}_2\text{O}$~\citep{zhang2023ho} selects important tokens based on attention scores. During the prefilling phase, approaches like SnapKV~\citep{li2024snapkv}, PyramidKV~\citep{zhang2024pyramidkv}, and LazyLLM~\citep{fu2024lazyllmdynamictokenpruning} cache only key input tokens, while MInference~\citep{jiang2024minference} and RetrievalAttention~\citep{liu2024retrievalattentionacceleratinglongcontextllm} use sparse attention to reduce latency. Some methods also directly compress input prompts~\citep{li-etal-2023-compressing,jiang-etal-2024-longllmlingua,pan-etal-2024-llmlingua}. Although these techniques compress the KV cache, they risk discarding critical tokens needed for later generation, which can lead to performance degradation~\citep{tang2024quest}.

\paragraph{Hidden State Reduction and Quantization}
Some other methods reduce hidden state size or quantize model weights. MQA~\citep{shazeer2019fast} and GQA~\citep{ainslie-etal-2023-gqa} group multiple heads into one, and MLA~\citep{liu2024deepseek} uses low-rank representations. AWQ~\citep{MLSYS2024_42a452cb} and QLLM~\citep{liu2024qllm} quantize weights and activations to save memory and computation.

\paragraph{Layer Redundancy Reduction}
Another line of work reduces redundancy between layers by sharing key-value states, as in LCKV~\citep{wu-tu-2024-layer}, CLA~\citep{brandon2024reducing}, and MiniCache~\citep{liu2024minicache}. Compared to these, our method: 1) leverages attention score similarity between layers and scales this to LLMs, and 2) adopts a head-wise sharing strategy based on a search process, rather than sharing only between adjacent or final layers.

\section{Conclusion}
\label{sec:conclusion}
We present \textsc{PoD}, a novel KV cache compression method for long-context LLMs. Unlike previous approaches that discard less important tokens, \textsc{PoD} retains all tokens and allocates memory based on token importance, using inter-layer attention redundancy to share key states for distant tokens. Experiments show that \textsc{PoD} reduces KV cache memory by up to 35\% with no loss in model performance. Our method is robust, generalizes well, and can be combined with token-selection techniques for further efficiency. We believe \textsc{PoD} offers a practical and effective solution for memory-efficient long-context inference in LLMs.

\bibliography{aaai2026}
\newpage
\appendix
\onecolumn
\section*{Impacts and Limitations}
\textsc{PoD} enables large language models to process much longer contexts with significantly reduced memory usage, without loss of performance. This makes long-context LLM deployment more practical in real-world scenarios and lowers hardware requirements for inference. The method can also be applied to various models and combined with other memory-saving techniques, supporting more efficient large-scale language model applications.

\textsc{PoD} requires a small amount of post-training to help the model adapt to the new attention-sharing structure. While this process is lightweight compared to full model training, it still introduces extra computational cost, especially for very large models. In addition, although \textsc{PoD} achieves strong memory savings without sacrificing accuracy, some latency overhead remains compared to the fastest token-eviction approaches, since all tokens are retained. Finally, the combination of \textsc{PoD} with other compression or quantization techniques has not been fully explored and may present new challenges or opportunities for further efficiency improvements.

\section{Baseline Details}
\label{sec:appendix.baseline}
\subsection{SnapKV}
For SnapKV, we followed the official implementation\footnote{\url{https://github.com/FasterDecoding/SnapKV}}, setting the \texttt{window\_size} to $4096$ and using the default value $64$ for \texttt{snap\_kv\_window\_size}. Additionally, it is worth noting that we extended SnapKV to support GQA~\citep{ainslie-etal-2023-gqa}, enabling its combination with \textsc{PoD}.

\subsection{PyramidKV}
For PyramidKV, we also used the official implementation\footnote{\url{https://github.com/Zefan-Cai/KVCache-Factory}}. The settings are: \texttt{window\_sizes} = $8$, \texttt{max\_capacity\_prompts} = $2048$, \texttt{kernel\_sizes} = $7$, \texttt{pooling} = `maxpool', and \texttt{window\_size} = $4096$.

\subsection{Quest}
For Quest, we consistently used the official codebase\footnote{\url{https://github.com/mit-han-lab/Quest}}, with \texttt{token\_budget} set to $4096$ and \texttt{chunk\_size} set to $16$.

\subsection{WA and WA+CPT}
For window attention, we set the \texttt{window\_size} parameter in FlashAttention\footnote{\url{https://github.com/Dao-AILab/flash-attention}} to (4096, 0).

\subsection{StreamingLLM and LM-Infinite}
Both StreamingLLM and LM-Infinite are based on the official StreamingLLM implementation\footnote{\url{https://github.com/mit-han-lab/streaming-llm}}, with the position embedding adapted for each method. We set \texttt{start\_size} to $16$ and \texttt{recent\_size} to $4080$.

\subsection{$\text{H}_2\text{O}$}
$\text{H}_2\text{O}$ is based on official implementation\footnote{\url{https://github.com/FMInference/H2O}}, with \texttt{start\_size} set to $96$ and \texttt{recent\_size} set to $4000$.

\subsection{CLA}
We re-implemented CLA~\citep{brandon2024reducing} and set the sharing factor to $2$.

\section{Additional Experiments}
\label{sec:obs}

\subsection{Window Size Impact on Prediction Consistency}
\label{subsec:obs1}

\begin{table}[h]
\centering

\renewcommand{\arraystretch}{1.3}
\setlength{\tabcolsep}{4mm}
\begin{tabular}{c|ccccc} 
\hline

\hline

\textbf{Window size} & $256$ & $512$ & $1024$ & $2048$ & $4096$ \\

\hline
\textbf{Identical rate} & $80.3$ & $82.2$ & $83.6$ & $84.8$ & $86.0$ \\
\hline

\hline
\end{tabular}

\caption{Proportion of identical predictions for the last 100 tokens under different recent window sizes.}
\label{tab:obs1}
\end{table}
To conduct this experiment, we sampled $1000$ texts of length $32$K from Dolma~\citep{soldaini-etal-2024-dolma}. All experiments were performed on our self-trained LLaMA3-8B-32K model. We evaluated the proportion of cases where the model predictions for the last $100$ tokens were exactly the same when using only a recent window of size \([256, 512, 1024, 2048, 2096]\), compared to using the full context. The detailed results are shown in Table~\ref{tab:obs1}.

\subsection{Offline Inter-Layer Attention Sharing Exploration}
\label{subsec:obs2}
\begin{figure}[h]
    \centering
    \includegraphics[width=\linewidth]{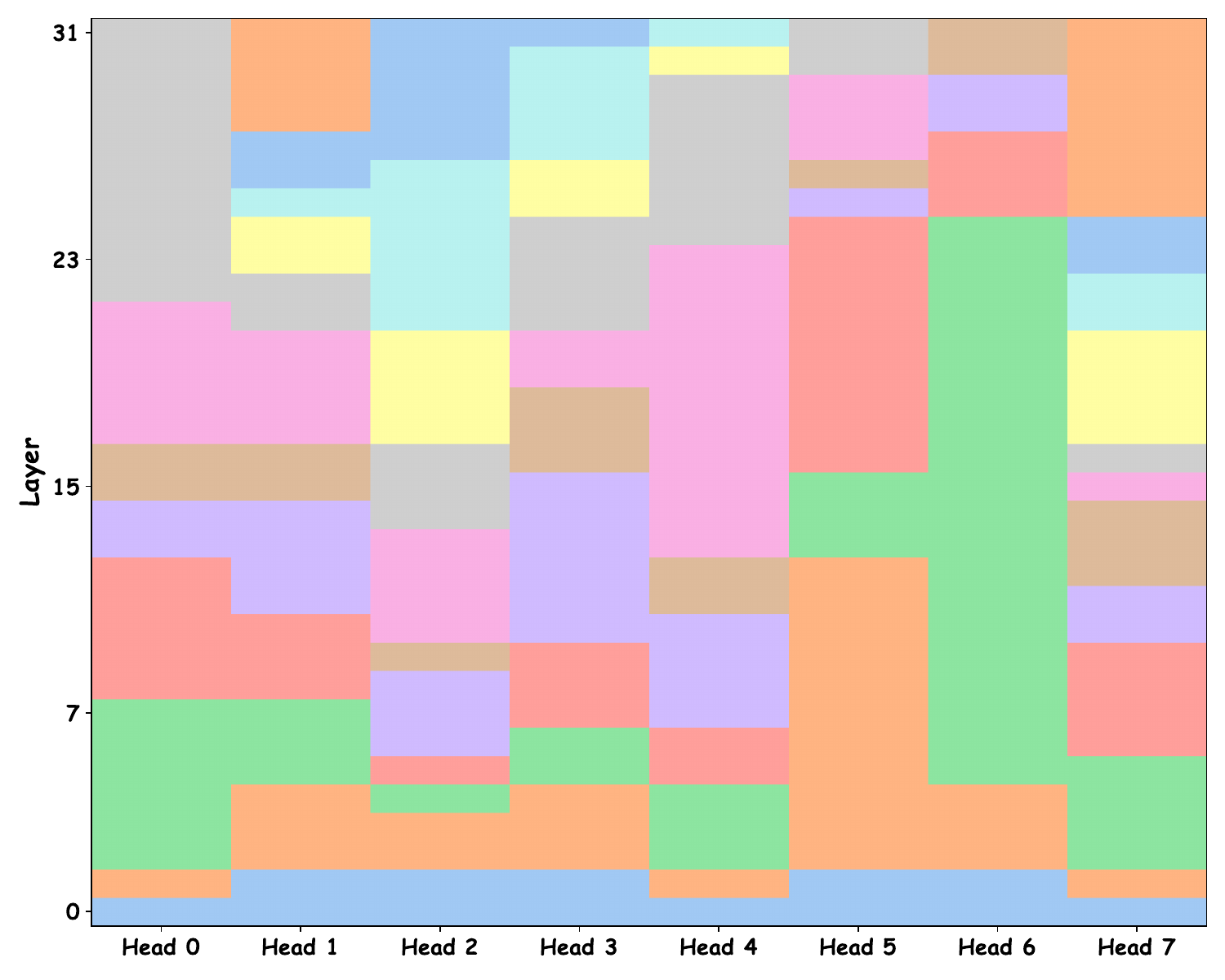}
    \caption{Offline exploration of inter-layer attention sharing for \textsc{PoD}. Each column corresponds to a head; consecutive layers with the same color within a head indicate a group of layers that share attention. Note that the same color in different heads does not imply any relationship between those layers.}
    \label{fig:offline}
\end{figure}

We sampled $1000$ sequences of length $32$K from Dolma~\citep{soldaini-etal-2024-dolma}, extracting the attention scores for the last $16$ tokens of each sequence. For GQA, the similarity between two layers for a group head is determined by a majority vote among the heads in that group. Figure~\ref{fig:offline} illustrates the layer sharing patterns of \textsc{PoD} described in \cref{subsec:exp.perf}.

\subsection{Visual Illustration of the Search Results for Needle in a Haystack}
\label{subsec:visrniah}

\begin{figure}[!h]
    \centering
    \includegraphics[width=\linewidth]{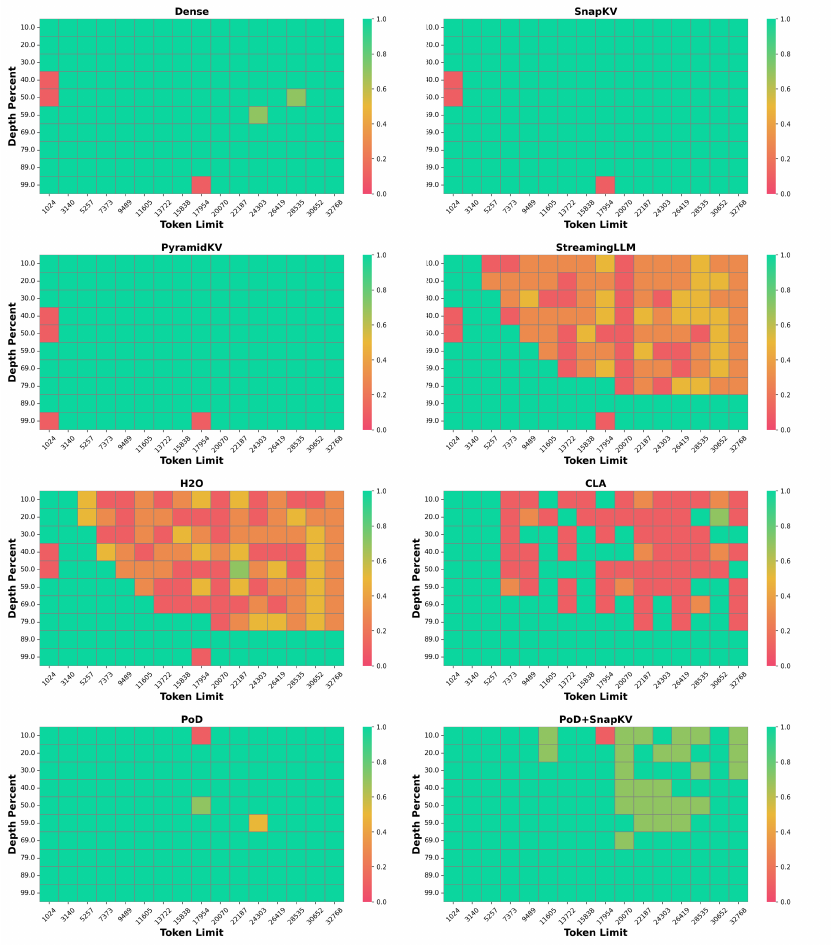}
    \caption{Visual Illustration of the Search Results for Needle in a Haystack}
    \label{fig:niah_viz}
\end{figure}

Figure~\ref{fig:niah_viz} provides a visual illustration of the search results for different methods. We observe that StreamingLLM and $\text{H}_2\text{O}$ fail to retrieve the needle when it falls outside their predefined window. In contrast, our method, which avoids token loss, performs comparably to dense models and is able to locate nearly all needles.

\subsection{Computation Optimization for Distant Tokens}
Empirical evidence suggests that in many situations, the prediction of the next token can be effectively accomplished without attending to distant tokens. This is reflected in Equation \ref{formula:gate}, where $g_{\ell, i}$ approaches $1$ in numerous cases. Based on this, for layers within a block that are not the lowest, we can preemptively evaluate the value of $g_{\ell, i}$. If $g_{\ell, i}\geq \tau$~($0\leq\tau\leq1$ is a hyperparameter), the computation of attention for distant tokens can be omitted, thereby reducing computation for distant tokens.

\begin{figure}[h]
    \centering
    \includegraphics[width=.7\linewidth]{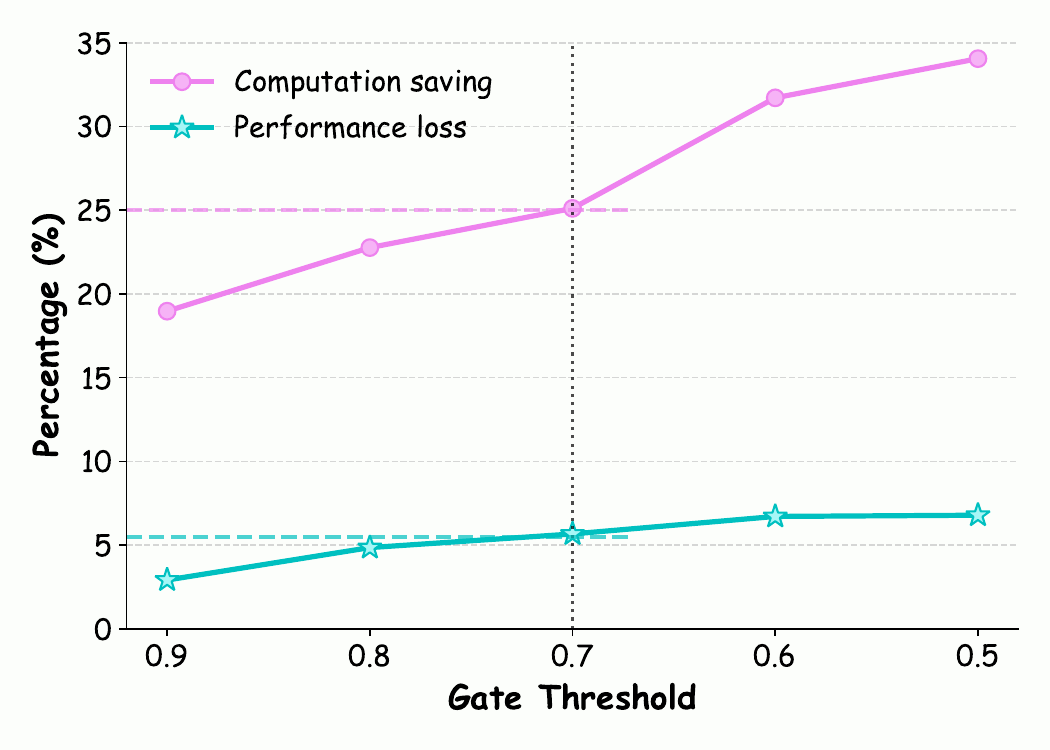}
    \caption{Computation saving and performance loss rates vs. the gate threshold $\tau$}
    \label{fig:gate}
\end{figure}

Figure \ref{fig:gate} shows the relationship between the ratio of computational savings and performance loss on LEval and the value of $\tau$. We observe that as $\tau$ decreases, it becomes easier to ignore the computation for distant tokens, leading to greater computational savings, but with some performance loss. However, when $\tau<0.7$, the performance degradation slows down while the computational savings become more pronounced. With $\tau=0.7$, computational cost is reduced by $25\%$ with only a $5\%$ performance drop.

\section{Theoretical Derivation}
\subsection{Derivation of Integrating Attention to Proximal and Distant Tokens}
\label{subsec:appendix_derivation}
For token $x_i$ at the $\ell$-th layer, we divide its context tokens into two groups: proximal tokens $T_P=\left\{j\mid x_j\text{ is a proximal token}\right\}$ and distant tokens $T_D=\left\{j\mid x_j\text{ is a distant token}\right\}$. The standard attention output to them is
\begin{equation}
    \begin{aligned}
        \mathbf{o}_{\ell, i} &= \frac{\sum\limits_{j\in T_P\cup T_D}\exp{\mathbf{a}_{\ell, i}^j}\cdot\mathbf{V}_{\ell, j}}{\sum\limits_{j\in T_P\cup T_D}\exp{\mathbf{a}_{\ell, i}^j}} \\
        &= \frac{\sum\limits_{j\in T_P}\exp{\mathbf{a}_{\ell, i}^j}\cdot\mathbf{V}_{\ell, j}}{\sum\limits_{j\in T_P\cup T_D}\exp{\mathbf{a}_{\ell, i}^j}} + \frac{\sum\limits_{j\in T_D}\exp{\mathbf{a}_{\ell, i}^j}\cdot\mathbf{V}_{\ell, j}}{\sum\limits_{j\in T_P\cup T_D}\exp{\mathbf{a}_{\ell, i}^j}} \\
        &= \frac{\sum\limits_{j\in T_P}\exp{\mathbf{a}_{\ell, i}^j}}{\sum\limits_{j\in T_P\cup T_D}\exp{\mathbf{a}_{\ell, i}^j}} \cdot \frac{\sum\limits_{j\in T_P}\exp{\mathbf{a}_{\ell, i}^j}\cdot\mathbf{V}_{\ell, j}}{\sum\limits_{j\in T_P}\exp{\mathbf{a}_{\ell, i}^j}} + \frac{\sum\limits_{j\in T_D}\exp{\mathbf{a}_{\ell, i}^j}}{\sum\limits_{j\in T_P\cup T_D}\exp{\mathbf{a}_{\ell, i}^j}} \cdot \frac{\sum\limits_{j\in T_D}\exp{\mathbf{a}_{\ell, i}^j}\cdot\mathbf{V}_{\ell, j}}{\sum\limits_{j\in T_D}\exp{\mathbf{a}_{\ell, i}^j}} \\
        &= \frac{\sum\limits_{j\in T_P}\exp{\mathbf{a}_{\ell, i}^j}}{\sum\limits_{j\in T_P\cup T_D}\exp{\mathbf{a}_{\ell, i}^j}} \cdot \mathbf{o}_{\ell, i}^P + \frac{\sum\limits_{j\in T_D}\exp{\mathbf{a}_{\ell, i}^j}}{\sum\limits_{j\in T_P\cup T_D}\exp{\mathbf{a}_{\ell, i}^j}} \cdot \mathbf{o}_{\ell, i}^D. \\
    \end{aligned}
\end{equation}
Therefore, we set
\begin{align}
    g_{\ell, i} &= \frac{\sum\exp{\mathbf{a}_{\ell, i}^P}}{\sum\exp{\mathbf{a}_{\ell, i}^P} + \sum\exp{\mathbf{a}_{\ell, i}^D}}.
\end{align}

\newpage

\end{document}